\pdfoutput=1
\documentclass[11pt]{article}

\usepackage[]{emnlp2021}
\usepackage{times}
\usepackage{latexsym}
\usepackage{svg}

\usepackage[russian,main=english]{babel}

\usepackage{hyphenat}

\usepackage{microtype}
\usepackage{times}
\usepackage{booktabs}
\usepackage{url}
\usepackage{latexsym}
\usepackage{multirow}
\usepackage{graphicx}
\usepackage{graphics}
\usepackage{chngpage}
\usepackage{caption}
\usepackage{subcaption}
\usepackage{pifont}
\usepackage{verbatim}

\title{Evaluating Multiway Multilingual NMT in the Turkic Languages}

\author{ \bf
Jamshidbek Mirzakhalov$^{a,b}$,  
Anoop Babu$^{a,b}$, 
Aigiz Kunafin$^a$,   
Ahsan Wahab$^a$, 
Behzod Moydinboyev$^{a,b}$,\\
\bf 
Sardana Ivanova$^{a,c}$,
Mokhiyakhon Uzokova$^{a,d}$,
Shaxnoza Pulatova$^{a,e}$,
Duygu Ataman$^{a,f}$, \\
\bf
Julia Kreutzer$^{a,g}$,
Francis Tyers$^{a,h}$,
Orhan Firat$^{a,g}$,
John Licato$^{a,b}$,
Sriram Chellappan$^{a,b}$
\\
\\

$^a$Turkic Interlingua, $^b$University of South Florida, \\ $^c$University of Helsinki,
$^d$Tashkent State University of Uzbek Language and Literature, \\ $^e$Namangan State University, $^f$NYU, $^g$Google Research,  $^h$Indiana University \\

}

\begin{document}
\maketitle
\begin{abstract}

Despite the increasing number of large and comprehensive machine translation (MT) systems, evaluation of these methods in various languages has been restrained by the lack of high-quality parallel corpora as well as engagement with the people that speak these languages. In this study, we present an evaluation of state-of-the-art approaches to training and evaluating MT systems in 22 languages from the Turkic language family, most of which being extremely under-explored \cite{joshi2019unsung}. First, we adopt the TIL Corpus \cite{mirzakhalov2021largescale} with a few key improvements to the training and the evaluation sets. Then, we train 26 bilingual baselines as well as a multi-way neural MT (MNMT) model using the corpus and perform an extensive analysis using automatic metrics as well as human evaluations. We find that the MNMT model outperforms almost all bilingual baselines in the out-of-domain test sets and finetuning the model on a downstream task of a single pair also results in a huge performance boost in both low- and high-resource scenarios. Our attentive analysis of evaluation criteria for MT models in Turkic languages also points to the necessity for further research in this direction. We release the corpus splits, test sets as well as models to the public\footnote{\url{https://github.com/turkic-interlingua/til-mt}}. 
\end{abstract}

\begin{figure}[]

\includegraphics[width=0.5\textwidth]{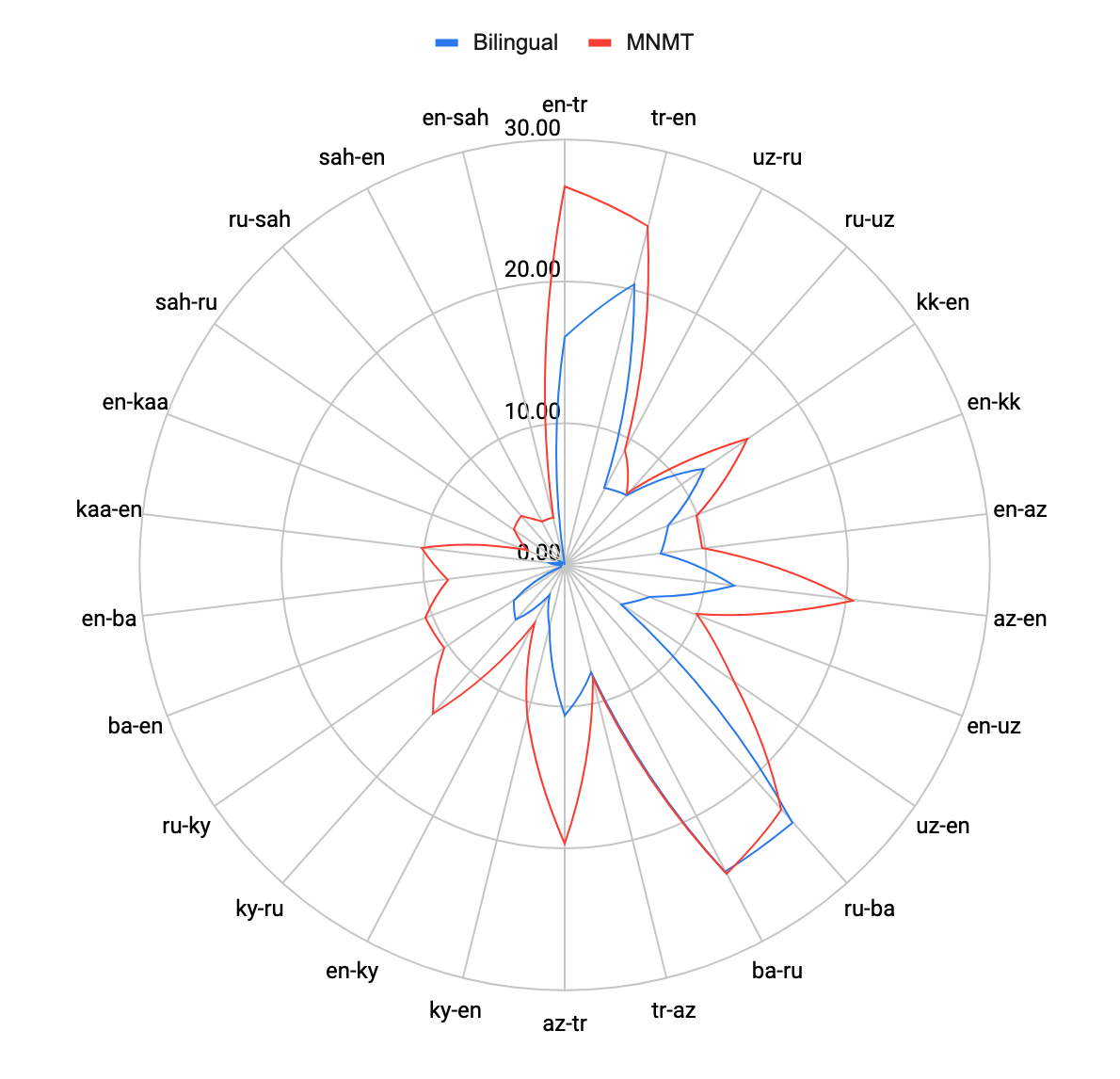}
\caption{Performance comparison between bilingual baselines and the MNMT model on X-WMT test set.}
\label{out-domain}
\end{figure}

\begin{table}[t]
\centering

\resizebox{0.48\textwidth}{!}{%
\begin{tabular}{llrrc}
\toprule
\textbf{Name}   & \textbf{Codes} & \textbf{Speakers} & \textbf{Data} & \textbf{MT?} \\ \midrule
English         & en, eng & 400.0M & 38.6M  & \ding{51}  \\
Russian         & ru, rus & 258.0M & 23.3M  & \ding{51}  \\
\midrule
Turkish         & tr, tur & 85.0M  & 52.6M  & \ding{51}  \\
Kazakh          & kk, kaz & 13.2M  & 5.3M   & \ding{51}  \\
Uzbek           & uz, uzb & 27.0M  & 2.9M   & \ding{51}  \\
Azerbaijani     & az, aze & 23.0M  & 2.2M   & \ding{51}  \\
Tatar           & tt, tat & 5.2M   & 1.8M   & \ding{51}  \\
Kyrgyz         & ky, kir & 4.3M   & 1.7M   & \ding{51}  \\
Chuvash         & cv, chv & 1.0M   & 1.5M   & \ding{51}  \\
Turkmen         & tk, tuk & 6.7M   & 910.4K & \ding{51}  \\
Bashkir         & ba, bak & 1.4M   & 880.5K & \ding{51}  \\
Uyghur          & ug, uig & 10.0M  & 334.8K & \ding{51}  \\
Karakalpak      & kaa     & 583.0K & 253.8K & \ding{55} \\
Khakas          & kjh     & 43.0K  & 219.0K & \ding{55} \\
Altai           & alt     & 56.0K  & 192.6K & \ding{55} \\
Crimean Tatar   & crh     & 540.0K & 185.3K & \ding{55} \\
Karachay-Balkar & krc     & 310.0K & 162.8K & \ding{55} \\
Gagauz          & gag     & 148.0K & 157.4K & \ding{55} \\
Sakha           & sah     & 450.0K & 157.1K & \ding{51}  \\
Kumyk           & kum     & 450.0K & 156.8K & \ding{55} \\
Tuvinian        & tyv     & 280.0K & 100.3K & \ding{55} \\
Shor            & cjs     & 3.0K   & 2.3K   & \ding{55} \\
Salar           & slr     & 70.0K  & 766    & \ding{55} \\
Urum            & uum     & 190.0K & 491    & \ding{55} \\ \bottomrule
\end{tabular}%
}
\caption{(The table indicates the language codes used for the Turkic languages along with the number of L1 speakers, amount of available data (in sentences) in our corpus. The column MT? indicates if there are currently available online machine translation systems for the language. K: thousand, M: million.)}
\label{tab:lang_stats}
\end{table}

\section{Introduction}
The last few years have seen encouraging advances in low-resource MT development with the increasing availability of public multilingual corpora~\citep{agic-vulic-2019-jw300,OrtizSuarezSagotRomary2019,ccmatrix,el-kishky-etal-2020-ccaligned,tiedemann2020tatoeba,goyal-etal-2021-flores-101,nekoto2020participatory} and more inclusive multilingual MT models~\citep{arivazhagan2019massively,TiedemannThottingal:EAMT2020,m2m-100}. In this study, we take the Turkic language family into focus, which has not been studied at large in MT research (detailed review in Section~\ref{sec:related}). Most recently, in a wide evaluation of translation between hundreds of languages with a multilingual model (M2M-124) trained on large web-mined parallel data, translation into, from, and between Turkic languages was shown to be very challenging compared to other language families~\citep{goyal-etal-2021-flores-101}. With the promise of strong transfer capabilities of multilingual models especially for related languages, we hope that the inclusion of a wider set of Turkic languages into a joint model can unlock automatic translation even for the very low-resourced Turkic languages where no prior translation models exist \cite{koehn2005europarl, choudhary2011creating, post2012constructing, nomoto2018tufs, espla2019paracrawl, nekoto2020participatory}.


To this aim, we adopt the TIL Corpus \cite{mirzakhalov2021largescale} compiled by the Turkic Interlingua\footnote{\url{https://turkicinterlingua.org/}} community \cite{mirzakhalov2021turkic} including X-WMT test sets with a few key improvements (Section~\ref{sec:corpus}). We train a multi-way NMT (MNMT) model on the entire parallel corpus, which constitutes the first large-scale multilingual translation model specifically for Turkic languages (Section~\ref{sec:experiments}). We perform an extensive analysis of the strengths and weaknesses of this model, comparing it to the bilingual baselines and evaluating it under a domain shift. We find that the MNMT model outperforms almost all bilingual baselines in the out-of-domain tests while it performs comparably or underperforms in the in-domain tests. We further analyze its capacity for transfer learning by fine-tuning the model on several language pairs all of which experience gains, both in- and out-of-domain scenarios. In addition, we complement the automatic evaluation with a human evaluation study for multiple languages (Section~\ref{sec:evaluation}), gaining insights into types of common mistakes that the model makes and the suitability of different automatic metrics for Turkic languages. We plan on releasing the improved corpus, evaluation sets, and all the models to the public.

This work will not only enrich the landscape of languages currently considered in MT research and spur future research on NLP for Turkic languages but will hopefully also inspire the building of new translation engines and derived technologies for populations with millions of native speakers (Table~\ref{tab:lang_stats}). 


\section{Related Work}\label{sec:related}

This section discusses the previous work on MT of these languages including the available corpora and languages resources. The 19 Turkic languages covered in the study are: Altai, Azerbaijani, Bashkir, Crimean Tatar, Chuvash, Gagauz, Karachay-Balkar, Karakalpak, Khakas, Kazakh, Kumyk, Kyrgyz, Sakha, Turkmen, Turkish, Tatar, Tuvan, Uyghur, and Uzbek. There are several other widely spoken languages that are left out from our study such as Shor, Salar, Urum, Nogai, Khorasani Turkic, Qashqai, and Khalaj, due to the lack (or very limited amount) of any available parallel corpora. Future work will focus on extending the corpus to these languages as well. 

\subsection{MT of Turkic Languages}

The need for more comprehensive and diverse multilingual parallel corpora has sped up the creation of such large-scale resources for many language families and linguistic regions \cite{koehn2005europarl, choudhary2011creating, post2012constructing, nomoto2018tufs, espla2019paracrawl, nekoto2020participatory}. \citet{tiedemann2020tatoeba} released a large-scale corpus for over 500 languages covering thousands of translation directions. The corpus currently includes 14 Turkic languages and provides bilingual baselines for all translation directions present in the corpus. However, most of the 14 Turkic languages contain a few hundred or a dozen samples. In addition, the varying and limited size of the test sets does not allow for the extensive analysis and comparisons between different model artifacts, linguistic features, and translation domains. More recently, \citet{goyal-etal-2021-flores-101} extended the previous Flores benchmark by providing human translated evaluation sets for 101 languages, among which 5 of them are from the Turkic family: Azerbaijani, Kazakh, Kyrgyz, Turkish, and Uzbek. Similarly, they train a large MNMT model and evaluate its performance using the benchmark.

A Russian-Turkic parallel corpus was curated for 6 different Turkic languages, and their bilingual baselines have been reported for both directions using different NMT-based approaches \citet{khusainov2020first}. However, the dataset, test sets, and models are not released to the public which limits its use to serve as a comparable benchmark. Additionally, a rule-based MT framework for Turkic languages has been presented with 4 language pairs \citet{alkim2019machine}. Also, several rule-based MT systems have been built for Turkic languages which are publicly available through the Apertium\footnote{\url{https://www.apertium.org/}} website \citet{apertium:2019}. 

For individual languages in our corpus, there are several proposed MT systems and linguistic resources: Azerbaijani \cite{hamzaoglu1993machine, fatullayev2008dilmanc}, Bashkir \cite {tyers2012prototype}, Crimean Tatar \cite {gokirmak2019machine, altintacs2001turkish}, Karakalpak \cite {kadirov2015algorithm}, Kazakh \cite{assylbekov2014initial, sundetova2015free, littell2019multi, briakou2019university, tukeyev2019neural}, Kyrgyz \cite{ccetin7assisting}, Sakha \cite{ivanova-etal-2019-tools}, Turkmen \cite {tantuug2018machine}, Turkish \cite{turhan1997english,el2006initial,bisazza2009morphological,tantuug2011turkmenceden, ataman2017linguistically}, Tatar \cite{salimzyanov2013free, khusainov2018building, valeev2019application, gokirmak2019machine}, Tuvan \cite{killackey2013statistical}, Uyghur \cite{mahsut2004experiment, 6311137,  song2015construction, wang2020njunlp}, and Uzbek \cite {axmedova2019algorithm}. Yet to our knowledge, there has not been a study that covers Turkic languages to such a large extent as ours, both in terms of multilingual parallel corpora and multiway NMT benchmarks across these languages.

\section{TIL Corpus}\label{sec:corpus}

As we adopt the TIL Corpus as the training data, we perform a few key modifications to better the quality of the datasets. 

First, we notice that the alignments for the Bible\footnote{\url{https://bible.is/}} and TedTalks\footnote{\url{https://www.ted.com/participate/translate}} datasets were not optimal as most "sentences" were actually comprised of multiple sentences in order to preserve the quality of the alignment with target sequence. For example, in the case of TedTalks, the original speech utterance may have been 2-3 sentences in text but the translation of that speech may end up differing by 1 or even more sentences depending on the translator. Common practice in this situation, as seen through multiple corpora across OPUS\footnote{\url{https://opus.nlpl.eu/}}, is to leave the entire utterance as is to preserve the quality of the alignment even if the number of sentences do not match. Instead, we drop the examples where the total number of sentences do not match and split (and realign) the cases where they do. This naturally increased the overall number of sentence alignments in both the Bible and TedTalks corpora for all language pairs. 

Second, we perform a corpus-wide length and length-ration filtering where we drop sentence pairs that are single words as well as the entries where source and target ratio is over 2. 

Third, we re-curate the in-domain evaluation sets following the improvements to the corpus. Details on the evaluation sets are described further in Section~\ref{evaluation}.

\subsection{Curation of evaluation sets}
\label{evaluation}
The original TIL Corpus introduced three evaluation sets with different domains (Bible, TedTalks, and X-WMT). To simplify the analysis of the models, we re-curate the in-domain evaluation sets by randomly sampling from each corpora. X-WMT is used as the out-of-domain test set since it is from the news domain with substantial amount of new words/terms that most of the language pairs lack. The curation steps for the test sets are presented below.

\subsubsection{In-domain Evaluation Sets}
In-domain development and test sets are randomly sampled from each language pair and can serve as evaluation sets for both bilingual and multilingual models. The size of the development and test sets depends on the amount of training data available. More specifically, development and test sizes are 5k each if the train size is over 1 million parallel sentences, 2.5k if over 100k, 1k if over 10k, and 500 if over 2.5k. All test and development samples are removed from the training corpus for that language pair. Overall, this yields development and test sets for exactly 400 language pairs. 

\subsubsection{X-WMT Test Set}
\label{evaluation:x-wmt}
X-WMT is a challenging and human-translated test set in the news domain based on the professionally translated test sets in English-Russian from the WMT 2020 Shared Task \citep{mathur-etal-2020-results}. It was originally introduced in the TIL Corpus and we adopt the test sets as they are. Currently, the test set extends into 8 Turkic languages (Bashkir, Uzbek, Turkish, Kazakh, Kyrgyz, Azerbaijani, Karakalpak, and Sakha) paired with English and Russian. Table~\ref{tab:wmt} highlights the currently available test set directions. Bolded entries in the table indicate the original direction of the translation.

\begin{table}[]
\centering
\resizebox{\columnwidth}{!}{%
\begin{tabular}{lrrrrrrrrrr}
\toprule
 & \textbf{en} & \textbf{ru} & \textbf{ba} & \textbf{tr} & \textbf{uz} & \textbf{ky} & \textbf{kk} & \textbf{az} & \textbf{sah} & \textbf{kaa} \\ 
 \midrule
\textbf{en}  & ---    &     &   &  &  &  &  &  &  &  \\ 
\textbf{ru}  & \textbf{1000}  &  ---   &   &  &  &  &  &  &  &  \\  
\textbf{ba}  & 1000  &  \textbf{1000}   & --- &   &  &  &  &  &  &  \\  
\textbf{tr}  & \textbf{800} & 800 & 800  &  --- &  &  &  &  &  &  \\  
\textbf{uz}  & \textbf{900} & 900 & 900 & 600 &  --- &  &  &  &  &  \\  
\textbf{ky}  & 500 & \textbf{500} & 500 & 400 & 500 & ---&  &  &  &  \\ 
\textbf{kk}  & 700 & 700 & 700 & 500 & \textbf{700} & 500 & ---&  & & \\  
\textbf{az}  & \textbf{600} & 600 &  600 & 500  & 600  & 500  & 500 & --- &  &   \\  
\textbf{sah} & 300 & \textbf{300} & 300 & 300 & 300 & 300 & 300 & 300 &  ---&   \\ 
\textbf{kaa} & 300 & 300 &  300  & \textbf{300}   &  300  & 300 & 300 & 300 & 300 & ---\\
\bottomrule
\end{tabular}%
}
\caption{X-WMT test sets. Bolded entries indicate the original translation direction.}
\label{tab:wmt}
\end{table}

\section{Experimental Setup}\label{sec:experiments}

\subsection{Bilingual Experiments}
To serve as initial baselines, we train 26 bilingual baselines using the corpus and report the performance on the in-domain test set as well as the X-WMT set (out-of-domain) as described in Section \ref{evaluation:x-wmt}. The selection of the language pairs was constricted by the availability of both in-domain and out-of-domain test sets to enable more meaningful insights from the experiments.

\subsubsection{Model details}
All models are Transformers (\textit{transformer-base}) \cite{vaswani2017attention} and are trained using the JoeyNMT framework \cite{kreutzer-etal-2019-joey}. In the preprocessing stage, we use Sacremoses for tokenization and apply byte pair encoding (BPE) \cite{sennrich2015neural, dong-etal-2015-multi} with a joint vocabulary size of 4k and 32k. Models use 512-dimensional word embeddings and hidden layers and are trained with the Adam optimizer \cite{kingma2015adam}. A learning rate of $3*10^{-4}$ is applied along with a dropout rate of 0.3. We use a batch size of 4096 BPE tokens with 8 accumulations to simulate training on 8 GPU machines. All models, except English-Turkish and Turkish-English, are trained on Google Colab's freely availably preemptible GPUs.

\subsection{Multilingual Experiments}


To examine the extent of transfer learning and generalization within our corpus, we train a multiway multilingual NMT model on the entire dataset covering almost 400 language directions. We then compare the performance of the model on the in-domain and out-of-domain test sets across a range of language pairs.

\begin{table*}[hbt!]
\centering
\resizebox{\textwidth}{!}{%
\begin{tabular}{lllllll|lllll}
\hline
       &                                 & \multicolumn{5}{c|}{\textbf{In-Domain Test}}                           & \multicolumn{5}{c}{\textbf{X-WMT Test}}                               \\ \hline
\textbf{Pairs } & \multicolumn{1}{l|}{\textbf{Train size}} & \multicolumn{2}{c}{\textbf{Bilingual}}     & \multicolumn{3}{c|}{\textbf{MNMT}} & \multicolumn{2}{c}{\textbf{Bilingual}}     & \multicolumn{3}{c}{\textbf{MNMT}} \\ \cline{3-12} 
 &
  \multicolumn{1}{l|}{} &
  \multicolumn{1}{c}{\textbf{BLEU}} &
  \multicolumn{1}{c|}{\textbf{Chrf}} &
  \multicolumn{1}{c}{\textbf{BLEU}} &
  \multicolumn{1}{c}{\textbf{Chrf}} &
  \multicolumn{1}{c|}{\textbf{PPL}} &
  \multicolumn{1}{c}{\textbf{BLEU}} &
  \multicolumn{1}{c|}{\textbf{Chrf}} &
  \multicolumn{1}{c}{\textbf{BLEU}} &
  \multicolumn{1}{c}{\textbf{Chrf}} &
  \multicolumn{1}{c}{\textbf{PPL}} \\ \hline
en-tr  & \multicolumn{1}{l|}{35.8M}      & 31.45 & \multicolumn{1}{l|}{0.51} & \textbf{33.09}   & \textbf{0.51}   & 8.18   & 16.04 & \multicolumn{1}{l|}{0.55} & \textbf{26.74}  & \textbf{0.56}  & 12.76   \\
tr-en  & \multicolumn{1}{l|}{35.8M}      & 31.37 & \multicolumn{1}{l|}{0.50} & \textbf{35.48}   & \textbf{0.52}   & 7.19   & 20.39 & \multicolumn{1}{l|}{0.51} & \textbf{24.66}  & \textbf{0.55}  & 10.88   \\
ru-uz  & \multicolumn{1}{l|}{1.3M}       & \textbf{53.12} & \multicolumn{1}{l|}{\textbf{0.76}} & 44.73   & 0.71   & 3.02   & 6.58  & \multicolumn{1}{l|}{0.41} & \textbf{6.70}   & \textbf{0.42}  & 82.20   \\
uz-ru  & \multicolumn{1}{l|}{1.3M}       & \textbf{55.39} & \multicolumn{1}{l|}{\textbf{0.76}} & 46.42   & 0.71   & 3.27   & 6.08  & \multicolumn{1}{l|}{0.36} & \textbf{9.16 }  & \textbf{0.39}  & 16.70   \\
en-kk  & \multicolumn{1}{l|}{564.8K}     & \textbf{24.53} & \multicolumn{1}{l|}{\textbf{0.54}} & 18.92   & 0.49   & 10.45  & 7.82  & \multicolumn{1}{l|}{0.40} & \textbf{9.92}   & \textbf{0.43}  & 10.02   \\
kk-en  & \multicolumn{1}{l|}{564.8K}     & \textbf{29.17} & \multicolumn{1}{l|}{\textbf{0.51}} & 24.67   & 0.48   & 7.47   & 12.00 & \multicolumn{1}{l|}{0.42} & \textbf{15.71}  & \textbf{0.44}  & 26.02   \\
az-en  & \multicolumn{1}{l|}{548.9K}     & \textbf{26.65} & \multicolumn{1}{l|}{\textbf{0.48}} & 20.47   & 0.42   & 7.70   & 12.01 & \multicolumn{1}{l|}{0.41} & \textbf{20.41}  & \textbf{0.49 } & 14.46   \\
en-az  & \multicolumn{1}{l|}{548.9K}     & \textbf{34.73} & \multicolumn{1}{l|}{\textbf{0.56}} & 15.27   & 0.42   & 8.74   & 6.79  & \multicolumn{1}{l|}{0.38} & \textbf{9.71}   & \textbf{0.43}  & 10.59   \\
en-uz  & \multicolumn{1}{l|}{529.6K}     & \textbf{45.95} & \multicolumn{1}{l|}{\textbf{0.66}} & 27.80   & 0.51   & 6.04   & 6.34  & \multicolumn{1}{l|}{0.40} & \textbf{9.89}   & \textbf{0.42}  & 47.45   \\
uz-en  & \multicolumn{1}{l|}{529.6K}     & \textbf{38.72} & \multicolumn{1}{l|}{\textbf{0.58}} & 32.44   & 0.50   & 6.15   & 4.81  & \multicolumn{1}{l|}{0.24} & \textbf{14.45}  & \textbf{0.45}  & 19.08   \\
ba-ru  & \multicolumn{1}{l|}{523.7K}     & \textbf{46.02} & \multicolumn{1}{l|}{\textbf{0.69}} & 40.59   & 0.64   & 3.75   & 24.39 & \multicolumn{1}{l|}{\textbf{0.58}} & \textbf{24.57}  & 0.57  & 5.49    \\
ru-ba  & \multicolumn{1}{l|}{523.7K}     & \textbf{51.26} & \multicolumn{1}{l|}{\textbf{0.74}} & 43.44   & 0.67   & 3.24   & \textbf{24.31} & \multicolumn{1}{l|}{\textbf{0.59}} & 23.13  & 0.56  & 6.29    \\
az-tr  & \multicolumn{1}{l|}{410.1K}     & \textbf{23.47} & \multicolumn{1}{l|}{\textbf{0.48}} & 18.40   & 0.43   & 8.87   & 10.61 & \multicolumn{1}{l|}{0.43} & \textbf{19.63}  & \textbf{0.48}  & 23.42   \\
tr-az  & \multicolumn{1}{l|}{410.1K}     & \textbf{29.97} & \multicolumn{1}{l|}{\textbf{0.53}} & 15.71   & 0.42   & 8.37   & 7.78  & \multicolumn{1}{l|}{0.39} & \textbf{8.21}   & \textbf{0.42}  & 14.51   \\
en-ky  & \multicolumn{1}{l|}{312.6K}     & \textbf{21.66} & \multicolumn{1}{l|}{\textbf{0.44}} & 14.54   & 0.38   & 10.77  & 2.33  & \multicolumn{1}{l|}{0.27} & \textbf{4.64}   & \textbf{0.34}  & 19.57   \\
ky-en  & \multicolumn{1}{l|}{312.6K}     & \textbf{24.96} & \multicolumn{1}{l|}{\textbf{0.42}} & 18.01   & 0.38   & 11.02  & 4.65  & \multicolumn{1}{l|}{0.29} & \textbf{10.87}  & \textbf{0.39}  & 35.64   \\
ky-ru  & \multicolumn{1}{l|}{293.7K}     & \textbf{19.63} & \multicolumn{1}{l|}{\textbf{0.40}} & 16.30   & 0.38   & 10.04  & 5.23  & \multicolumn{1}{l|}{0.30} & \textbf{14.08}  & \textbf{0.44}  & 9.43    \\
ru-ky  & \multicolumn{1}{l|}{293.7K}     & \textbf{18.57} & \multicolumn{1}{l|}{\textbf{0.43}} & 14.82   & 0.40   & 9.58   & 4.42  & \multicolumn{1}{l|}{0.35} & \textbf{10.35}  & \textbf{0.45}  & 11.52   \\
ba-en  & \multicolumn{1}{l|}{34.3K}      & \textbf{21.51} & \multicolumn{1}{l|}{0.36} & 17.79   & \textbf{0.37}   & 10.81  & 0.32  & \multicolumn{1}{l|}{0.19} & \textbf{10.55}  & \textbf{0.40}  & 37.89   \\
en-ba  & \multicolumn{1}{l|}{34.3K}      & \textbf{17.78} & \multicolumn{1}{l|}{0.33} & 17.29   &\textbf{0.35}   & 10.52  & 0.16  & \multicolumn{1}{l|}{0.14} & \textbf{8.35}   & \textbf{0.34}  & 21.43   \\
en-kaa & \multicolumn{1}{l|}{17.1K}      & 15.34 & \multicolumn{1}{l|}{0.40} & \textbf{19.42}   & \textbf{0.46}   & 8.83   & 0.31  & \multicolumn{1}{l|}{0.19} & \textbf{2.82}   & \textbf{0.27}  & 77.93   \\
kaa-en & \multicolumn{1}{l|}{17.1K}      & \textbf{22.82} & \multicolumn{1}{l|}{0.43} & 21.95   & \textbf{0.48}   & 8.56   & 1.04  & \multicolumn{1}{l|}{0.21} & \textbf{10.21}  & \textbf{0.38}  & 38.17   \\
ru-sah & \multicolumn{1}{l|}{9.2K}       & \textbf{13.26} & \multicolumn{1}{l|}{\textbf{0.35}} & 5.46    & 0.19   & 30.82  & 0.12  & \multicolumn{1}{l|}{0.16} & \textbf{4.64}   & \textbf{0.17}  & 58.01   \\
sah-ru & \multicolumn{1}{l|}{9.2K}       & \textbf{16.35} & \multicolumn{1}{l|}{\textbf{0.36}} & 13.11   & 0.26   & 23.00  & 0.42  & \multicolumn{1}{l|}{0.18} & \textbf{4.41}   & \textbf{0.25}  & 40.68   \\
en-sah & \multicolumn{1}{l|}{8.1K}       & \textbf{13.45} & \multicolumn{1}{l|}{\textbf{0.36}} & 4.98    & 0.18   & 34.31  & 0.04  & \multicolumn{1}{l|}{\textbf{0.14}} & \textbf{3.46}   & 0.12  & 75.38   \\
sah-en & \multicolumn{1}{l|}{8.1K}       & \textbf{22.19} & \multicolumn{1}{l|}{\textbf{0.40}} & 5.90    & 0.23   & 24.58  & 0.16  & \multicolumn{1}{l|}{0.21} & \textbf{3.38}   & \textbf{0.24}  & 110.50  \\ \hline
\end{tabular}%
}
\caption{Experiments results from bilingual baselines and MNMT model evaluated on the in-domain and out-of-domain test sets. \textit{BLEU} and \textit{Chrf} uses the SacreBLEU implementation and \textit{PPL} refers to the internal perplexity of the MNMT model.}
\label{tab:in-out}
\end{table*}

\begin{figure}[]

\includegraphics[width=0.5\textwidth]{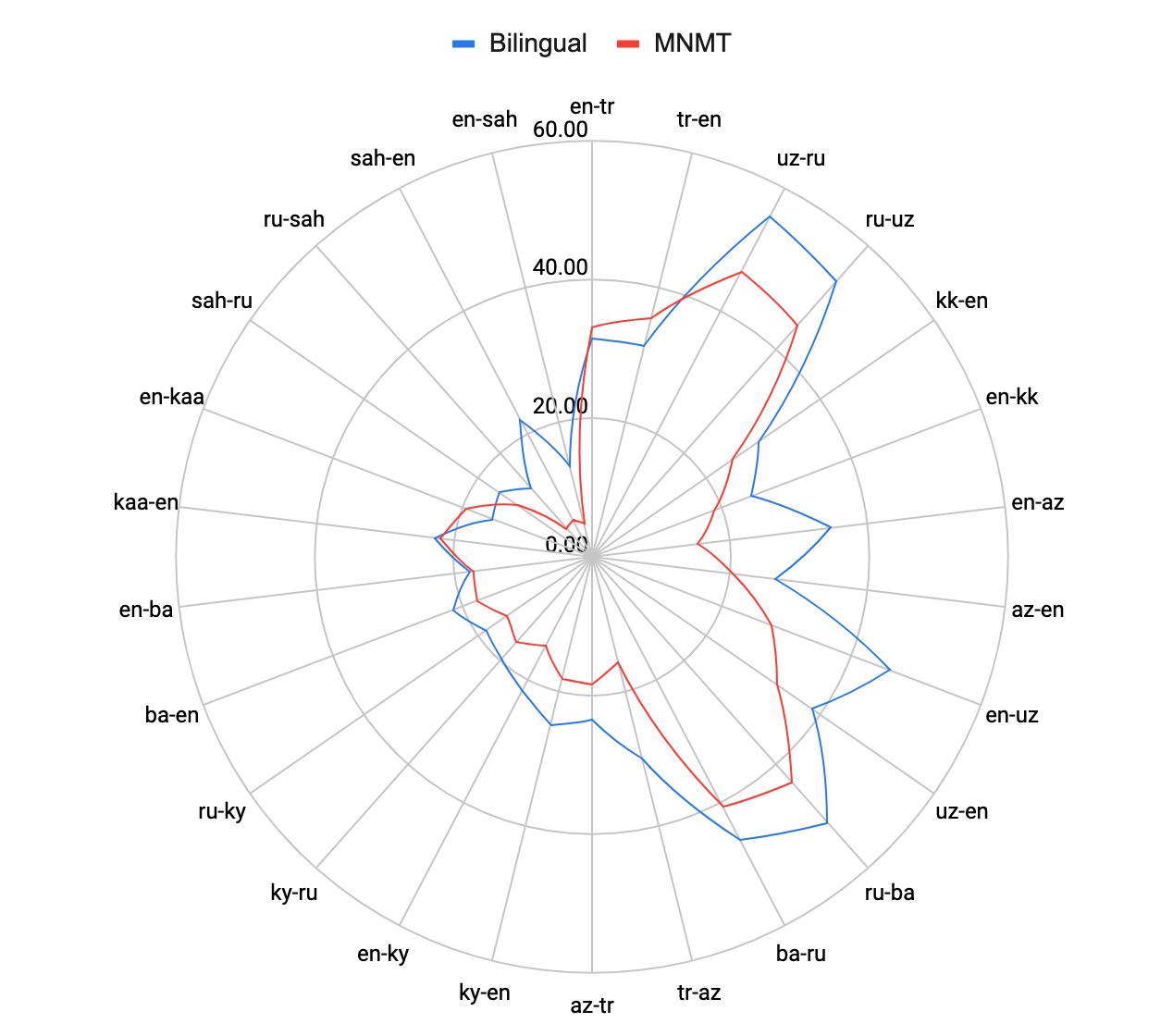}
\caption{Performance comparison between bilingual baselines and the MNMT model on the in-domain test set.}
\label{in-domain}
\end{figure}

\subsubsection{Data Preprocessing}
Similar to the bilingual data preprocessing, the entire corpus has been tokenized using Sacremoses\footnote{\href{https://github.com/alvations/sacremoses}{https://github.com/alvations/sacremoses}} and samples longer than 300 words have been filtered out. In addition, we perform cross-filtering of test and dev sets of all language pairs from the training corpus, as it is very necessary to do so in any MNMT model using a multiway corpus. Since the corpus is relatively unbalanced, we perform a temperature-based sampling with a value of 1.25. Although a higher temperature value between 2 and 3 would further balance our corpus, it would increase the dataset size by 8x with t=2 and 25x with t=3. This increase would limit our ability to train the model due to the restrained compute resources. Originally, the overall training set size is at around 133 million samples and this increases to 244 million after the sampling procedure. We apply the sentencepiece\footnote{\href{https://github.com/google/sentencepiece}{https://github.com/google/sentencepiece}} implementation of the byte pair encoding (BPE)~\citep{sennrich-etal-2016-neural} with a joint vocabulary size of 64k. Following the method from \citet{ha2016toward}
and \citet{johnson2017google}, we prepend a target language token to the source sentences to enable many-to-many translation.


\subsubsection{Model details}
We train the model using the Transformer architecture in the \textit{transformer-base} configuration. More specifically, we use the \textit{transformer\_wmt\_en\_de} version from Fairseq~\citep{ott2019fairseq} implementation\footnote{\url{https://github.com/pytorch/fairseq/tree/master/examples/translation}} with 6 layers both in the encoder and decoder. Configuration of the model closely follows the original implementation of the Transformer~\citep{vaswani2017all} with the model dimension set at 512 and hidden dimension size at 2048. We apply a dropout rate of 0.3, the learning rate of $5*10^{-4}$, and warm-up updates of 40k. The effective batch size is 16,384 BPE tokens. The model is trained using 4 NVIDIA V100 GPU machines for a little over 1 million steps which takes about 36-48 hours.

\subsection{Evaluation of Models}
Automatic evaluation metrics used to compare the performance of bilingual baselines and MNMT are token-based corpus BLEU~\citep{papineni-etal-2002-bleu} and character-based Chrf~\citep{popovic-2015-chrf}. While corpus BLEU is the de-facto standard in MT~\citep{marie-etal-2021-scientific}, Chrf might work better for morphologically rich languages because it can reward partially correct words. We also report the MNMT model's internal perplexity to better highlight the language pairs in which the model struggles most. 
We evaluate the models on the in-domain and X-WMT evaluation sets. The gap between scores on in-domain versus out-of-domain translations is particularly interesting since it gives us an estimate of domain robustness and generalization, as well as mimics a realistic shift from the training domain to the domain of interest for potential users or downstream applications.



\begin{figure*}[hbt!]
\centering
\includegraphics[width=0.8\textwidth]{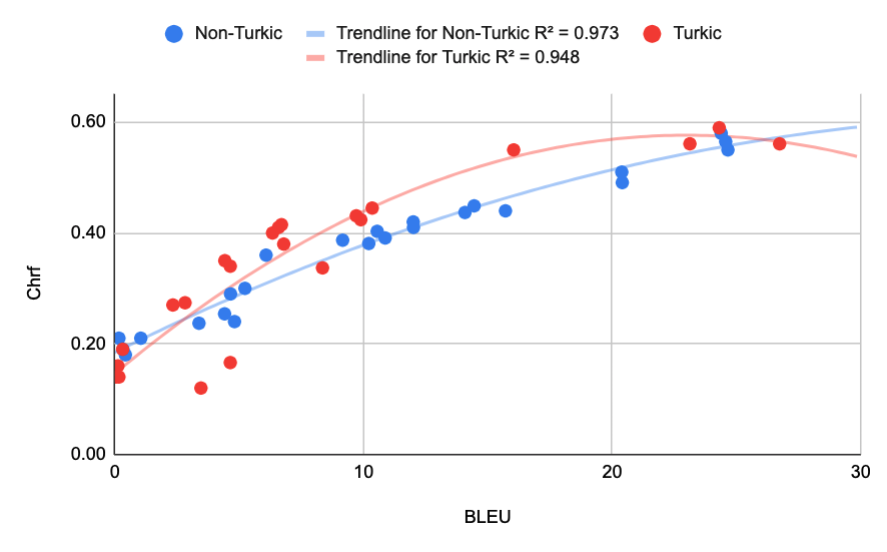}
\caption{Correlations between BLEU and Chrf scores when the target language is Turkic and non-Turkic.}
\label{chrf_bleu_comp}
\end{figure*}

\begin{table}[]
\centering
\resizebox{\columnwidth}{!}{%
\begin{tabular}{lll|ll|ll}
\hline
                          & \multicolumn{2}{c|}{\textbf{Bilingual}} & \multicolumn{2}{c|}{\textbf{MNMT}} & \multicolumn{2}{c}{\textbf{Gain}} \\ \hline
\multicolumn{1}{l|}{} &
  \multicolumn{1}{c}{\textbf{BLEU}} &
  \multicolumn{1}{c|}{\textbf{Chrf}} &
  \multicolumn{1}{c}{\textbf{BLEU}} &
  \multicolumn{1}{c|}{\textbf{Chrf}} &
  \multicolumn{1}{c}{\textbf{BLEU}} &
  \multicolumn{1}{c}{\textbf{Chrf}} \\
\multicolumn{1}{l|}{XX-En} & 6.92               & 0.31               & 13.78            & 0.42            & +6.86           & +0.11           \\
\multicolumn{1}{l|}{En-XX} & 4.57               & 0.30               & 9.37             & 0.36            & +4.80           & +0.06           \\
\multicolumn{1}{l|}{XX-Ru} & 9.03               & 0.36               & 13.06            & 0.41            & +4.03           & +0.06           \\
\multicolumn{1}{l|}{Ru-XX} & 8.86               & 0.38               & 11.21            & 0.40            & +2.35           & +0.02           \\
\multicolumn{1}{l|}{XX-XX}  & 8.99               & 0.38               & 12.49            & 0.41            & +3.49           & +0.04           \\ \hline
\end{tabular}%
}
\caption{Performance comparison with different language groups and their overall gains in the MNMT setup. XX refers to the Turkic languages in the corpus.}
\label{tab:lang-family}
\end{table}

\subsection{Bilingual baselines vs MNMT}

Table~\ref{tab:in-out} shows all the results for the bilingual baselines and MNMT as evaluated on two test tests. The first obvious trend in the table is the dominance of the bilingual baselines on the in-domain test sets as they overperform the MNMT model in most of the high- to mid-resource language pairs. As the train size decreases, the results become more comparable in terms of BLEU and even better for MNMT when evaluated in Chrf. When tested under a domain shift with the X-WMT set, MNMT results in gains across almost all pairs. However, it is important to note that there is a noticeable performance drop that follows the domain shift as can be seen in Figures~\ref{out-domain} and~\ref{in-domain}. This highlights a realistic phenomenon of generalization and sets an expectation of the model's capabilities in real-world use cases. 

Another observation in Table~\ref{tab:in-out} is that all of the language pairs having fewer than 100k training samples (8 total) in our bilingual baselines barely pass the mark of 1 BLEU score or 0.2 Chrf in the out-of-domain test. However, in the MNMT setup, the average BLEU and Chrf score for those 8 low-resource pairs are 5.98 and 0.27 respectively. While these scores indicate that these pairs are still extremely low in quality and potentially unusable in practice, gains are promising given the amount of resources and a moderately-sized MNMT model. 

To examine the generalization of the MNMT model into different language groups, we calculate the average gains for all pairs translating into English (XX-En), from English (En-XX), into Russian (XX-Ru), from Russian (Ru-XX), and direct pairs (XX-XX). Table~\ref{tab:lang-family} shows the average gains per category in terms of BLEU and Chrf. As it looks, translating from and into English sees the most gains, which is very consistent with the findings from the community \cite{arivazhagan2019massively, goyal-etal-2021-flores-101}. A positive trend is the increasing quality of direct pairs which are very comparable to the non-Turkic pairs. We hypothesize that one of the main reasons for this is that the TIL Corpus is a multi-centric dataset with training data between almost all language pairs which allows us to train a complete Multilingual Neural Machine Translation (cMNMT) \cite{freitag2020complete}. As shown in \cite{freitag2020complete, fan2021beyond}, MNMT models trained on multi-centric parallel corpora tend to result in performance gains between non-English pairs.

\begin{table}[]
\centering
\resizebox{0.5\textwidth}{!}{%
\begin{tabular}{l|rrrc|rrrr}
\hline
               & \multicolumn{4}{c|}{\textbf{Adequacy}} & \multicolumn{4}{c}{\textbf{Fluency}} \\ \hline
 &
  \multicolumn{1}{l}{\textbf{Avg}} &
  \multicolumn{1}{c}{\textbf{k}} &
  \multicolumn{1}{c}{\textbf{LL}} &
  \textbf{UL} &
  \multicolumn{1}{l}{\textbf{Avg}} &
  \multicolumn{1}{c}{\textbf{k}} &
  \multicolumn{1}{c}{\textbf{LL}} &
  \multicolumn{1}{c}{\textbf{UL}} \\
\textbf{en-tr} & 2.97     & 0.33    & 0.23    & 0.43    & 3.20    & 0.12    & 0.04    & 0.21   \\
\textbf{tr-en} & 2.95     & 0.45    & 0.36    & 0.55    & 3.18    & 0.40    & 0.30    & 0.50   \\
\textbf{en-uz} & 2.77     & 0.18    & 0.10    & 0.26    & 2.93    & 0.28    & 0.17    & 0.38   \\
\textbf{uz-en} & 3.05     & 0.28    & 0.20    & 0.37    & 3.19    & 0.29    & 0.18    & 0.39   \\
\textbf{ba-ru} & 2.74     & 0.58    & 0.48    & 0.67    & 3.34    & 0.63    & 0.54    & 0.73   \\
\textbf{ru-ba} & 2.81     & 0.27    & 0.17    & 0.37    & 3.06    & 0.19    & 0.09    & 0.29   \\ \hline
\end{tabular}%
}
\caption{\textbf{Avg} represents the average score for either Adequacy or Fluency given by the annotators for each language pair. \textbf{k} represents the Cohen's Kappa score. \textbf{LL} represents the Lower Limit within 95\% confidence. \textbf{UL} represents the Upper Limit within 95\% confidence.}
\label{tab:kapp_stats}
\end{table}

\subsection{BLEU vs Chrf}
Figure~\ref{chrf_bleu_comp} compares BLEU and Chrf for all bilingual and multilingual models on X-WMT. We distinguish between translating into and from Turkic languages since all Turkic languages feature agglutination. As hinted above, we suspect that BLEU might underestimate translation quality when translating into Turkic languages. The graph shows a clear distinction that confirms this: For translations into non-Turkic languages, the relation between Chrf and BLEU is almost linear, with a Pearson correlation of 0.98 and a rank correlation of 0.98 as well.
For translation into Turkic, the trend follows a more curved line, with a largely higher Chrf-to-BLEU ratio. The Pearson correlation is much lower at 0.87, but the rank correlation is only slightly lower than for non-Turkic languages at 0.92. Consequently, we can expect the same BLEU score to correspond to a higher Chrf score when translating into Turkic languages than from them. This means that while Chrf and BLEU are likely to produce similar rankings of systems (at least in our scenario with standard comparable Transformer models), the Chrf score might better characterize the absolute translation quality. Our human evaluation does not cover sufficient language pairs (three from and three into Turkic languages) to yield a reliable empirical confirmation for this hypothesis. Future studies of larger scale as in the WMT metrics shared task~\citep{mathur-etal-2020-results} will be needed.

\begin{table*}[htp]
    \centering
    \resizebox{\textwidth}{!}{%
    \begin{tabular}{|p{0.5\linewidth}  p{0.5\linewidth}|}
    
    \hline
    	\multicolumn{2}{|c|}{\textbf{en-tr}} \\
    \hline
    	\multicolumn{2}{|l|}{\textbf{Adequacy:} 3.00 | \textbf{Fluency:} 4.00 |  \textit{Fluent Output with Inadequate Verbal Tense}} \\ \hline
        \texttt{Reference} Toyota, Subaru'daki hissesini 'den fazla artıracağını söyledi. &  
        \texttt{Hypothesis} Toyota, Subaru'daki hisseyi 'den fazla artırdığını söyledi. \\
     \hline
    	\multicolumn{2}{|l|}{\textbf{Adequacy:} 4.00 | \textbf{Fluency:} 3.00 |  \textit{Lexical choice preserves meaning, still not the natural construction}} \\ \hline
         \texttt{Reference} Başka birisi ağır yaralandı. &  
         
        \texttt{Hypothesis} Başka bir kişi kötü yaralandı. \\
         \hline

     \hline
    	\multicolumn{2}{|c|}{\textbf{tr-en}} \\
    	\hline
    	\multicolumn{2}{|l|}{\textbf{Adequacy:} 3.00 | \textbf{Fluency:} 2.00 |  \textit{Some of the translations made lost the original meaning}} \\ \hline
         \texttt{Reference} The schoolgirl who died from catastrophic injuries following a \textbf{suspected} hit-and-run in Newcastle has been \textbf{pictured} for the first time. &  
         
        \texttt{Hypothesis} After a \textbf{suspicious} hit-and-run in Newcastle's, the student who died badly was first \textbf{seen}. \\
    \hline
        \multicolumn{2}{|l|}{\textbf{Adequacy:} 3.00 | \textbf{Fluency:} 4.00 |  \textit{Maintains grammatical form, but changes the meaning}} \\ \hline
         \texttt{Reference} He further \textbf{dismissed} the embargo as an \textbf{attack} on the rights of citizens. &  
         
        \texttt{Hypothesis} He also \textbf{denied} the ambargo by defending an \textbf{attack} on citizens' rights. \\
    \hline
    
    \hline
    	\multicolumn{2}{|c|}{\textbf{ba-ru}} \\
    	\hline
    	\multicolumn{2}{|l|}{\textbf{Adequacy:} 2.00 | \textbf{Fluency:} 3.00 |  \textit{"kiss" translates to "kill" and changes the meaning completely}} \\ \hline
    	
        \texttt{Reference} 
        \foreignlanguage{russian}{В ночь после выборов, пишет Ло Бьянко в своей книге, Карен Пенс отказалась \textbf{поцеловать} мужа.} & 
        \texttt{Hypothesis}
        \foreignlanguage{russian}{В свою книгу Ло Бьянко, в ночь после выборов, Карен Пенс отказывается от \textbf{смерти} мужа.} \\
    \hline
        \multicolumn{2}{|l|}{\textbf{Adequacy:} 3.00 | \textbf{Fluency:} 2.00 |  \textit{Incorrect pronoun ("she" to "he"). Few awkward translations}} \\ \hline
         \texttt{Reference} \foreignlanguage{russian}{Поэтому она откликнулась на вакансию в Fast Trak Management, маленькой компании, которая называет себя "маркетинговой фирмой номер один в Северной Вирджинии".} &  
         
        \texttt{Hypothesis} \foreignlanguage{russian}{Поэтому \textbf{он согласился} на вакансию Fast Trak Management \textbf{в малой} компании, которая называла себя "Первую маркетинговую фирму в Северной Вирджинии".} \\
    \hline
     
    \hline
    	\multicolumn{2}{|c|}{\textbf{ru-ba}} \\
    	\hline
    	\multicolumn{2}{|l|}{\textbf{Adequacy:} 3.00 | \textbf{Fluency:} 2.00 |  \textit{When several verbs are present, some are omitted from the translation}} \\ \hline
         \texttt{Reference} \foreignlanguage{russian}{Видео Пирзаданың бер нисә йылан һәм аллигаторҙо тотоп торғанын күрһәтә.} &  
         
        \texttt{Hypothesis} \foreignlanguage{russian}{Дәүләт еренә йәмәғәт} \textbf{access} \foreignlanguage{russian}{Видеоға ярашлы, Пирзада бер нисә йылан һәм алгигатор менән нисек эш итә.} \\
    \hline
    	\multicolumn{2}{|l|}{\textbf{Adequacy:} 2.00 | \textbf{Fluency:} 3.00 |  \textit{A whole part of the original sentence is omitted from the translation}} \\ \hline
         \texttt{Reference} \foreignlanguage{russian}{iHandy тарафынан киң билдәле эмодзи-ҡушымталар серияһы сығарылды, әммә улар ҙа Google Play Store системаһынан шунда уҡ юйылды.} &  
         
        \texttt{Hypothesis} \foreignlanguage{russian}{iHandy Google Play Store-ҙан сығарылған популяр эмодзи-приложениялар серияһы булдырылды.} \\
    \hline
     
    \hline
    	\multicolumn{2}{|c|}{\textbf{uz-en}} \\
    	\hline
    	\multicolumn{2}{|l|}{\textbf{Adequacy:} 2.00 | \textbf{Fluency:} 2.00 |  \textit{Changed the order events}} \\ \hline
    	
         \texttt{Reference} Antonio Brown has indicated he's not retiring from the NFL, only a few days after announcing he was done with the league in a rant.	 &  
         
        \texttt{Hypothesis} Antonio Braun said that after a few days after the NFL, he won’t leave after he announced that he was engaged in league. \\
    \hline
        \multicolumn{2}{|l|}{\textbf{Adequacy:} 2.00 | \textbf{Fluency:} 4.00 |  \textit{Improper changes from original nouns, and different sense of "hold"}} \\ \hline
         \texttt{Reference} \textbf{Harker} says \textbf{Fed} should '\textbf{hold} firm' on interest rates &  
         
        \texttt{Hypothesis} \textbf{Everyone} thinks that this is how to \textbf{hold} the \textbf{Federal} rate percentages. \\
    \hline

    \hline
    	\multicolumn{2}{|c|}{\textbf{en-uz}} \\
    	\hline
    	\multicolumn{2}{|l|}{\textbf{Adequacy:} 3.00 | \textbf{Fluency:} 3.00 |  \textit{"Gumonlanuvchi":"a suspect"."Shubhachi":"someone who suspects"}} \\ \hline
         \texttt{Reference} Keyin ushbu mashinadan uch nafar \textbf{gumonlanuvchi} tushayotganini ko 'rishdi. &  
         
        \texttt{Hypothesis} Keyinchalik uchta \textbf{shubhachi} mashinadan chiqib ketganini ko'rishdi. \\
    \hline
        \multicolumn{2}{|l|}{\textbf{Adequacy:} 2.00 | \textbf{Fluency:} 2.00 |  \textit{Use of a correct but a foreign word (başarısız)}} \\ \hline
         \texttt{Reference} WeWork's Neumann muvaffaqiyatsiz IPO o 'tkazilgandan so'ng o 'zini bosh direktor lavozimidan chetlatishga ovoz berdi &  
         
        \texttt{Hypothesis} \textbf{Biz Work''s} Neumann IPO \textbf{başarısız} bo'lganidan so'ng \textbf{O'zbekiston} Bosh direktori sifatida ovoz berdi \\
    \hline
     
    \end{tabular}
    }
    \caption{Qualitative Analysis of the MNMT model output for 6 language pairs. The \texttt{Reference} sentence shows the intended translation while the \texttt{Hypothesis} shows the MNMT model output.}
    \label{tab:qualitative_smaples}
\end{table*}
\section{Human Evaluation of MNMT}\label{sec:evaluation}

\subsection{Human evaluation setup}
To facilitate analysis on how well evaluation metrics measure the quality of the translations, we conduct human evaluations using the outputs from the MNMT model on the X-WMT set. We use Direct Assessment (DA) and follow the TAUS guidelines\footnote{\href{https://rb.gy/eqlgbm}{https://rb.gy/eqlgbm}} with the only exception being the number of annotators per language pair, where we employ 2 annotators per language pair instead of 4\footnote{Due to limited resources.}. In our DA, two hundred sentences of the MNMT model's output per language pair are evaluated based on its adequacy and fluency on respective 1-4 point scales. Annotators received an explanation of the rating scales with the task (e.g. ``Adequacy: On a 4-point scale rate how much of the meaning is represented in the translation: 4: Everything 3: Most 2: Little 1: None''). To measure the inter-annotator agreement (IAA) between the two annotators of each language pair, we compute the Weighted Cohen's Kappa statistic \cite{doi:10.1177/001316446002000104}.


\begin{table*}[hbt!]
\centering
\resizebox{0.8\textwidth}{!}{%
\begin{tabular}{ll|ll|ll}
\hline
\multicolumn{1}{c}{\textbf{Pairs}} & \multicolumn{1}{c|}{\textbf{Train Size}} & \multicolumn{2}{c|}{\textbf{In-Domain}}                       & \multicolumn{2}{c}{\textbf{X-WMT}}                           \\ \hline
\multicolumn{1}{c}{}      & \multicolumn{1}{c|}{}           & \multicolumn{1}{c}{\textbf{BLEU}} & \multicolumn{1}{c|}{\textbf{Chrf}} & \multicolumn{1}{c}{\textbf{BLEU}} & \multicolumn{1}{c}{\textbf{Chrf}} \\
ru-ba  & 523.7K & 54.48 (+11.04) & 0.743 (+0.07) & 24.85 (+0.54)  & 0.569 (-0.02) \\
ky-en  & 312.6K & 24.21 (+6.2)   & 0.42 (+0.05)  & 10.26 (+5.61)  & 0.38 (+0.09)  \\
en-ba  & 34.3K  & 30.43 (+13.14) & 0.46 (+0.11)  & 4.56 (+4.4)    & 0.22 (+0.08)  \\
ru-sah & 9.2K   & 49.46 (+44.00) & 0.585 (0.4)   & 22.05 (+21.93) & 0.348 (+0.19) \\ \hline
\end{tabular}
}
\caption{Experiment results from the finetuning of the MNMT model.}
\label{tab:finetune}
\end{table*}


The language pairs involved in this human study are English-Turkish, Turkish-English, Bashkir-Russian, Russian-Bashkir, Uzbek-English, and English-Uzbek. These pairs were selected on the basis of language and script diversity, their performance on the X-WMT test set, and the availability of annotators.

\subsection{Discussion and Results}

The results of the average adequacy and fluency for each language pair are shown by Table \ref{tab:kapp_stats}. Most of the chosen language pairs received an average score of around 3 for both adequacy and fluency. This indicates that the model was largely able to convey most intended meaning in a good grammatical sense to a native speaker. Fluency is consistently rated higher than adequacy, which is a common theme in NMT evaluation~\citep{martindale-etal-2019-identifying}. The large difference in BLEU (5 BLEU points) between en-uz and uz-en is still noticeable, but much smaller according to the human evaluation. Chrf estimates a quality difference of 0.3 here, which is closer to the human estimate.

The Cohen's Kappa scores for each language pair are present in Table \ref{tab:kapp_stats}. As Cohen's Kappa is a measure from 0--1 of how well the two annotators agreed with their evaluations while removing possible agreements by chance, Cohen's Kappa score serves as one metric in deciding the reliability of the evaluations. We see that the reliability varies across language pairs and between adequacy and fluency. Translation into English or Russian has a higher agreement on average than in the opposite direction (en/tr is a tie).

\subsection{Qualitative Analysis}

To gain better qualitative insight into the model outputs in each of the 6 language directions, we asked the annotators to identify 2 examples that highlight the most commonly witnessed mistakes during their review. Table \ref{tab:qualitative_smaples} showcases those examples along with a brief explanation for their scores. From this analysis, it seems that the severity of mistakes that the MNMT model makes in \textit{adequacy} tends to range from certain words being translated to a slightly different meaning to the original intention of the sentence being lost. As for \textit{fluency}, the errors seem to range from awkward wording to clear grammatical mistakes. There are a few cases where there is an off-target translation for a word or a segment of the sentence.

\section{The Promise of MNMT: Cross-lingual Knowledge Transfer}

One of the biggest advantages of a large MNMT model is its capacity for transfer learning as can be accomplished through fine-tuning. Since we plan on releasing the model to the public, we believe many understudied and underperforming language pairs could benefit from cross-lingual knowledge transfer. This phenomenon is well-known in the broader NLP community as well as in MT research. To test this hypothesis, we fine-tune our MNMT model on 4 language pairs ranging from high(er)-resource to extremely low-resource in training data available. Table~\ref{tab:finetune} shows the results of the experiments. As it can be seen, the performance of the models improves steadily across all resource types, low-resource cases experiencing gains up to 44 BLEU points (or 0.4 Chrf) from the bilingual baselines in the in-domain evaluation. However, in out-of-domain scenarios, gains are not as significant. Mid- to high-resource pairs improve modestly in the range of 1--5 BLEU points (or 0--0.1 Chrf) while a low-resource pair, Russian-Sakha gains up to 22 BLEU points (0.19 Chrf). 

\section{Future Work and Conclusion}
In this work, we train and evaluate the first large-scale MNMT model for the Turkic language family which consists of many underexplored languages. Among many results, we find it very promising to train and finetune a MNMT model with a language family corpus as it boosts the cross-lingual knowledge transfer between the related languages and consistently improves over the strong bilingual baselines in out-of-domain scenarios. Our analysis also shows that Chrf and BLEU do not correlate in the same when the target language group if different: BLEU underestimates the translations for the Turkic languages. 

In the future work, we hope to include more of the underrepresented Turkic language pairs in the study and explore the potential of transfer learning into the translation of unseen languages and language pairs ("zero-shot").

\section*{Acknowledgements}

We thanks all of the members and partners of the Turkic Interlingua (TIL) community for their contributions to the project. Namely, we would like to thank our dedicated translators and annotators: Nurlan Maharramli, Sariya Kagarmanova, Iskander Shakirov, Aydos Muxammadiyarov, Ziyodabonu Qobiljon qizi, Alperen Cantez, Doniyorbek Rafikjonov, Mukhammadbektosh Khaydarov, Madina Zokirjonova, Erkinbek Vokhabov, Petr Popov, Abilxayr Zholdybai and Akylbek Khamitov. We also acknowledge and appreciate significant dataset contributions from Rasul Karimov, Khan Academy O'zbek\footnote{\url{https://uz.khanacademy.org/}}, and the Foundation for the Preservation and Development of the Bashkir Language\footnote{\url{https://bsfond.ru/}}.

\bibliography{anthology,custom, mirzakhalov}
\bibliographystyle{acl_natbib}




\end{document}